\documentclass[runningheads]{llncs}
\usepackage{graphicx}
\usepackage{algorithmic}
\usepackage{array}
\usepackage{mdwmath}
\usepackage{mdwtab}
\usepackage{eqparbox}
\usepackage{url}
\usepackage{amssymb}
\usepackage{lipsum}
\usepackage{stfloats}
\usepackage{amsmath}
\usepackage{subcaption}
\usepackage{multirow}

\newcommand{\transpose}{\mathsf{T}}

\makeatletter

\begin{document}
\title{DAEMA: Denoising Autoencoder with Mask Attention}
%
%
\author{Simon~Tihon\thanks{equal contribution}\orcidID{0000-0002-3985-1967}
\and Muhammad~Usama~Javaid\textsuperscript{\@fnsymbol{1}}\orcidID{0000-0001-9262-2250}
\and Damien~Fourure\orcidID{0000-0001-5085-0052}
\and Nicolas~Posocco\orcidID{0000-0002-1795-6039}
\and Thomas~Peel\orcidID{0000-0003-2967-6381}}

\institute{EURA NOVA, Mont-St-Guibert, Belgium\\
\email{firstname.lastname@euranova.eu}}
%
\authorrunning{S. Tihon et al.}
%
%

\maketitle              

\begin{abstract}
Missing data is a recurrent and challenging problem, especially when using machine learning algorithms for real-world applications. For this reason, missing data imputation has become an active research area, in which recent deep learning approaches have achieved state-of-the-art results. We propose DAEMA (\textit{Denoising Autoencoder with Mask Attention}), an algorithm based on a denoising autoencoder architecture with an attention mechanism.
While most imputation algorithms use incomplete inputs as they would use complete data - up to basic preprocessing (e.g. mean imputation) - DAEMA leverages a mask-based attention mechanism to focus on the observed values of its inputs.
We evaluate DAEMA both in terms of reconstruction capabilities and downstream prediction and show that it achieves superior performance to state-of-the-art algorithms on several publicly available real-world datasets under various missingness settings.
\end{abstract}


\begin{figure}[t!]
    \centering
    \includegraphics[width=\textwidth]{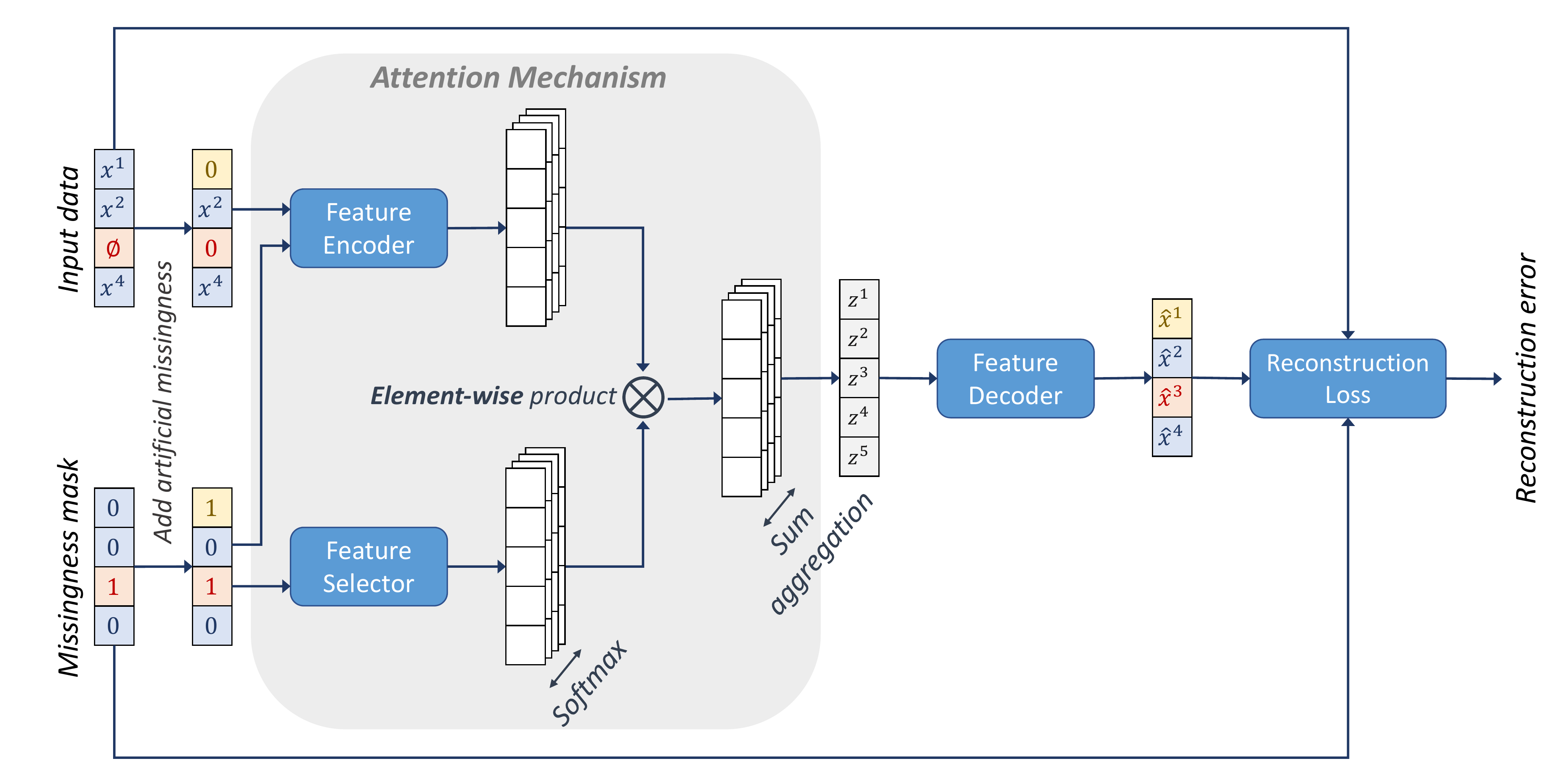}
    \caption{The goal of DAEMA is to produce a representation of the input which is robust to missingness, so that a \textit{Feature Decoder} can impute the missing data from it. To do so, a \textit{Feature Encoder} produces multiple values for each latent feature. These are then weighted by a \textit{Feature Selector} based on the missingness mask. This mask enables the attention mechanism to focus on the values that are produced using only observed inputs.
}
    \label{fig:full_arch}
\end{figure}

\section{Introduction}
Machine learning researchers and practitioners frequently encounter the problem of missing data. Data can be missing for many reasons. These include system failures, loss of data, or the fact that data was never known, measured or recorded.
Missing data can introduce bias and alter the statistical properties of a dataset. This can impact the performance of models learnt from this data, both in obvious (e.g. poor performance due to lack of data) or subtle (e.g. a bias is learnt) ways. Simple approaches consist in discarding samples with missing data, removing an entire feature if it is too often missing or imputing the mean/median value per feature. In practice, these provide a satisfying solution when a small proportion of the data is missing. Otherwise, it is likely to significantly alter the empirical data distribution.

Missing data imputation has received a lot of attention and many approaches have been proposed recently \cite{Aimnet,MCFlow,optimal_transport}. The literature can be divided into two categories: discriminative and generative approaches. Discriminative approaches model the conditional probability of each feature given the others. That is, they use all features but one to impute the remaining feature, and iterate over all features to produce a complete dataset. On the other hand, generative approaches model the joint distribution of all features to impute missing data all at once.

Generative approaches for missing data imputation generally involve some form of denoising autoencoder  \cite{DAE}.
A denoising autoencoder deals with noise, taking noisy samples as input and learning to reconstruct the cleaned samples. As missing data is a special case of noisy data, a denoising autoencoder can be used to reconstruct the missing parts.

\textit{Attention} is a very popular technique in the fields of natural language processing and computer vision. Introduced in \cite{attention-is-all}, attention mechanisms can be summed up as paying attention to parts of the input which are relevant to generate an output. Attention enables a model to understand the underlying structure of data better, resulting in better generalisation \cite{Aimnet}.

To leverage this promising technique, we propose DAEMA (\textit{Denoising Autoencoder with Mask Attention}). It improves on a simple yet efficient denoising autoencoder architecture by adding an attention mechanism based on the missingness mask. Its architecture can be seen in Figure \ref{fig:full_arch}. Thanks to a mask-based attention module, the latent representation produced by DAEMA is a missingness-robust embedding of the original sample. This approach achieves good performance with both randomly and systematically missing data, beating previous state-of-the-art techniques.


\section{Related Work}

Early approaches have modelled the task of missing data imputation as a predictive task. These are referred to as discriminative approaches. Some of them are based on k-nearest neighbours \cite{knn-imp} and support vector techniques \cite{svr-imp}, using models whose capacity is often insufficient for the task. Others are iterative, such as MissForest \cite{MissForest} and Multiple Imputation using Chained Equation (MICE) \cite{MICE}, which respectively use random forests and linear regressors to predict missing values for one feature at a time. Both approaches impute a same dataset several times, using the previous result to compute the next one more precisely. Since they produce multiple imputation of said dataset, these are referred to as multiple-imputation methods. Having multiple-imputed datasets can be an advantage as it helps to account for the uncertainty in the imputation process. However, these iterative methods are meant to impute whole datasets at once. This can be a drawback in real-world situations, where new data comes everyday.

Due to recent advances in deep learning and more particularly in generative deep-learning, more recent works have modelled the task of missing data imputation as a generative task. The resulting methods are referred to as generative approaches. One such method is called Multiple Imputation using Denoising Autoencoders (MIDA) \cite{MIDA}. It applies the denoising autoencoder approach to missing data imputation. Another method, called Generative Adversarial Imputation Nets (GAIN) \cite{GAIN}, leverages a conditional GAN \cite{GAN} to learn the real distribution of data through adversarial training. However, GANs are known to be hard to train, suffering from non-convergence and mode collapse. WGAN-GP \cite{wgan-gp} refines GANs to overcome these problems and ensure better training. Other methods, such as VAE \cite{VAE-imp} and HI-VAE \cite{DBLP:journals/HI-VAE}, use a variational autoencoder architecture. More recently, MCFlow \cite{MCFlow} leverages normalising flows and Monte Carlo sampling for imputation. Finally, GINN \cite{GINN} exploits graph convolutional networks to take advantage of the information contained in the nearest neighbours at imputation time instead of using only the information contained in each sample being imputed.

A recent deep-learning discriminative method, AimNet \cite{Aimnet}, achieves state-of-the-art results with a dot-product attention mechanism applied to the individual embedding of each feature. Its attention weights depend only on which feature is being imputed.

Finally, some variants of existing methods have been proposed. In \cite{IMDIDGM}, embeddings for categorical input features and gumbel-softmax activation layers for categorical output features are successfully applied to GAIN and to a VAE architecture. In \cite{optimal_transport}, the optimal transport distance between two batches is proposed to be used as a training loss for missing data imputation models. These approaches are complementary to our method.

We propose DAEMA, a generative method based on a denoising autoencoder and an attention mechanism. In contrast to AimNet, our attention mechanism is input-oriented: the attention weights of DAEMA depend on which feature values are missing instead of depending on which feature we want to impute. Therefore, the model can focus on non-missing values and distinguish them from placeholder values. To our knowledge, DAEMA is the first method to use an input-oriented attention mechanism for the imputation problem.


\section{Problem Statement}
\label{sec:problem_statement}

A dataset is defined as $D = \{\mathbf{X},\ C,\ N\}$. $\mathbf{X} \in \mathbb{R}^{n\times d}$ is a matrix of data composed of $n$ samples of $d$ features $\mathbf{x}_i = (x_i^1,\ \ldots,\ x_i^d) \in \mathbb{R}^d$. $\ C$ (resp. $N$) is the set of indices of categorical (resp. numerical) features, that is, features taking discrete (resp. continuous) values. In this work, in order to keep the loss and metrics simple, we focus on datasets containing only numerical features, i.e. $C = \emptyset$.

For the missing data imputation task, a dataset with missing values in $\mathbf{X}$ is given. We define the missingness-mask matrix $\mathbf{M} \in \{0, 1\}^{n\times d}$ such that $x_i^j$ is missing if and only if $m_i^j = 1$.
We denote $D^* = \{\mathbf{X^*},\ C,\ N\}$ the ground truth dataset without missing data.
Missing data are generally classified into three different categories \cite{rubin1976inference,Seaman2013WhatIM}:
\begin{itemize}
    \item Missing completely at random (MCAR) means the probability that a value is missing does not depend on any value in the dataset.
    \item Missing at random (MAR) means the probability that a value is missing depends only on the observed (non-missing) values.
    \item Missing not at random (MNAR) means the probability that a value is missing depends on unobserved values or latent variables.
\end{itemize}

We are interested in learning an imputation function
$
f: \mathbb{R}^d \times \{0,1\}^d \to \mathbb{R}^{d}; (\mathbf{x}, \mathbf{m}) \mapsto f(\mathbf{x}, \mathbf{m}).
$
Using the dataset $D$ and the corresponding missingness-mask matrix $\mathbf{M}$, the goal is to find the best function $f^{\dag}$ minimizing a reconstruction metric and a metric based on a downstream machine-learning task as described in Section~\ref{sec:Metrics}.


\section{Our Approach}
Figure \ref{fig:full_arch} shows the architecture of DAEMA, which is based on a standard denoising autoencoder. At its core is a mask-based attention mechanism, designed to help the network efficiently use the available data to produce robust latent representations. This in turn makes imputation possible by decoding these very representations.

\subsection{Denoising Autoencoder}
As a denoising autoencoder, DAEMA takes a noisy input sample $\mathbf{x}$ and produces a clean version $\mathbf{\hat{x}}$ of it. In our case, noise is defined as missingness of data, meaning that ground truth values of the missing data are unknown. It is trained using additional \textit{artificial missingness} in each batch, as the values of the missing features are needed to train the network. As it is impossible for the model to distinguish between originally missing values and artificially missing values, the model reconstructs both. Let $\mathbf{m}$ be the original missingness mask of the input sample $\mathbf{x}$, $\mathbf{\bar{m}}$ the one including artificial missingness and $\mathbf{\bar{x}} = \mathbf{x} \cdot (\mathbf{1}-\mathbf{\bar{m}})$ the sample with artificially missing values.

Because the originally missing values are unknown, the reconstruction loss has to take into account the missingness mask $\mathbf{m}$, as done in GAIN \cite{GAIN} and MCFlow \cite{MCFlow}. The masked reconstruction loss used is defined as follows:
\begin{equation}\label{eq:reconstruction_loss}
    \ell(\mathbf{\hat{x}}_i, \mathbf{x}_i, \mathbf{m}_i) = \sum_j (1 - m_i^j) \cdot (x_i^j - \hat{x}_i^j)^2
\end{equation}
Minimising the loss implies correctly imputing artificially missing data as well as correctly reconstructing observed values.
However, the minimisation of the loss is not impacted by originally missing data.

\subsection{Mask-Based Attention Mechanism}\label{subsub:attention}

To help the model focus on non-missing values, we add an attention mechanism into the encoder part of the network. The intuition is that, for different missingness patterns, the model has to focus on different non-missing values to reconstruct the missing ones. By adding an attention mechanism based on the missingness mask, the model can choose the values it has to focus on. As shown in Figure \ref{fig:full_arch}, the attention mechanism is based on three elements: the \textit{Feature Encoder}, the \textit{Feature Selector} and a \textit{Sum aggregation}.

The \textit{Feature Encoder} $f_e$ is a function that takes as input a sample with artificial missingness $\mathbf{\bar{x}}$ and its corresponding missingness mask $\mathbf{\bar{m}}$ and produces $d_z$ feature vectors of dimension $d'$: 
\begin{equation}\label{eq:fe}
    f_e(\mathbf{\bar{x}}, \mathbf{\bar{m}}) = (\mathbf{f}^1,\ldots, \mathbf{f}^{d_z}) \text{ with } \mathbf{f}^j \in \mathbb{R}^{d'}
\end{equation}
In practice, the \textit{Feature Encoder} is implemented as a multilayer perceptron producing an output of size $d' \cdot d_z$. This output is then reshaped into a two-dimensional feature map of size $(d', d_z)$.

The \textit{Feature Selector} $f_s$ is a function that takes the artificial missingness mask $\mathbf{\bar{m}}$ as input and produces $d_z$ selection vectors of dimension $d'$:
\begin{equation}\label{eq:fs}
    f_s(\mathbf{\bar{m}}) = (\mathbf{s}^1,\ldots, \mathbf{s}^{d_z}) \text{ with } \mathbf{s}^j \in \mathbb{R}^{d'}
\end{equation}
It is also implemented using a multi-layer perceptron which produces an output of size $d' \cdot d_z$. This output is then reshaped into a two-dimensional feature map of size $(d', d_z)$.

The objective of the \textit{Feature Encoder} is to create multiple estimation values for each latent feature (i.e. one feature vector $\mathbf{f}^j$ per latent feature $z^j \in \mathbf{z}$) while the goal of the \textit{Feature Selector} is to give more attention to the most meaningful values from each vector $\mathbf{f}^j$ according to the artificial missingness mask $\mathbf{\bar{m}}$.

The \textit{Attention Mechanism} combines feature vectors and selection vectors. To do so, the selection vectors $\mathbf{s}^j$ are normalized with a \textit{softmax} function $\sigma$. Then, an element-wise product is used to combine each feature vector $\mathbf{f}^j$ with its selection vector $\mathbf{s}^j$. Finally, a summation aggregates each resulting vector to get the final latent representation $\mathbf{z}$. Mathematically:
\begin{equation}
    \mathbf{z} = (\sigma(\mathbf{s}^1)^{\transpose} \cdot \mathbf{f}^1,\ldots,\sigma(\mathbf{s}^{d_z})^{\transpose} \cdot \mathbf{f}^{d_z}) = (z^1,\ldots, z^{d_z}).
\end{equation}

\section{Technical Implementation}
In order to ensure the reproducibility of our results, the architecture and implementation details are thoroughly defined in this section. 
The full training and testing code of DAEMA is available\footnote{\url{https://github.com/euranova/DAEMA}}.

\subsection{Preprocessing Steps}
\label{sec:preprocesing_steps}

To evaluate the imputation performance of DAEMA, complete datasets (i.e. without missing values) are needed. To train the model, we create real-world-like datasets $D$ (with missing values) from ground truth datasets $D^*$. The preparation of all datasets follows these steps:
\begin{itemize}
    \item We separate the downstream target label from the other features. In real-world applications, the target is unknown during the imputation process as it is predicted by the downstream model afterwards.
    \item For a few datasets, we remove some samples, such as samples containing naturally missing data or extreme outliers, as detailed in Section~\ref{sec:experimentation_datasets}.
    \item We introduce missing values in the ground truth datasets. We use the two \textit{uniform} mechanisms described in the MIDA paper \cite{MIDA} to create both MCAR and MNAR data. For the MCAR setting, each single value has 20\% chance to be removed, possibly removing none or all the values of a sample. For the MNAR setting, we randomly choose two features and select the samples for which a) the first feature has a value smaller or equal to the median of that first feature or b) the second feature has a value bigger or equal to the median of that second feature. Then, each single value of the selected samples has 20\% chance to be removed.
    \item We randomly split each dataset into a train set and a test set. We use a 70-30 ratio for all experiments.
    \item We apply a z-normalisation on the non-missing values by subtracting the mean and dividing by the variance of the training set.
\end{itemize}

\subsection{Detailed Architecture}

Inspired by AimNet \cite{Aimnet} and GAIN \cite{GAIN}, DAEMA has been kept as shallow as possible. 
The \textit{Feature Encoder} is a multilayer perceptron with two layers both using the hyperbolic tangent as activation. The input size is $2d$ ($\mathbf{\bar{x}}$ concatenated with $\mathbf{\bar{m}}$). Both hidden and output layers are of size $d' \cdot d_z$. The output of the \textit{Feature Encoder} is reshaped into a $(d', d_z)$ feature map.
The \textit{Feature Selector} is a single fully-connected layer without activation going from dimension $d$ to dimension $d' \cdot d_z$. The output vector is reshaped into a two dimensional matrix of size $(d', d_z)$.
The \textit{Feature Decoder} is a single fully-connected layer of dimension $d$ without activation.
For our experiments, $d'$ and $d_z$ are set to $2d$.

During the training, the artificial missingness mask $\mathbf{\bar{m}}$ and its corresponding sample $\mathbf{\bar{x}}$ are built on the fly by randomly removing features with a $0.2$ probability.
Besides, we use the masked reconstruction loss defined in equation \ref{eq:reconstruction_loss}. 
The model is trained by gradient descent using the Adam optimizer \cite{kingma2014adam} with a learning rate of $0.001$ and a batch-size of $64$. The training is stopped after $40,\!000$ batch-steps.

All hyperparameters are chosen based on early experiments. Fine-tuning these hyperparameters could probably improve the performance of the algorithm. However, this is out of the scope of this paper and is left for future work.


\section{Experimentation}
To validate our approach, we compare it to state-of-the-art algorithms on different datasets and with different missingness proportions.

\subsection{Datasets}
\label{sec:experimentation_datasets}
We use seven publicly available real-life datasets, six from the UCI repository \cite{UCI} and one from the \texttt{sklearn.datasets} module \cite{scikit_learn}: EEG Eye State (14,976 samples, 14 features, 2 classes), Glass (214 samples, 9 features, 6 classes), Breast Cancer (683 samples, 9 features, 2 classes), Ionoshpere (351 samples, 34 features, 2 classes), Shuttle (58,000 samples, 9 features, 7 classes),Boston Housing (506 samples, 13 features, regression) and CASP (45,730 samples, 9 features, regression). These datasets have different sizes and different dimensions to perform comprehensive experimentation. Note that the Breast Cancer dataset has samples with \textit{NA} values and the EEG dataset has four extreme outliers in its sixth feature. We remove these samples from the datasets. Boston, Glass, Breast Cancer, Ionosphere and Shuttle datasets are used for experimentation in MIDA~\cite{MIDA}. EEG and CASP datasets are used for experimentation in AimNet~\cite{Aimnet}. As these two models are state-of-the-art and direct competitors to our model, it is more relevant to perform comparison on these datasets.

\subsection{Metrics} \label{sec:Metrics}
We use two metrics to compare the algorithms, namely a reconstruction metric and a downstream metric. The reconstruction metric is the normalised root mean square error (\textit{NRMS}) over the missing values of the test set, as in \cite{Aimnet}. The normalisation is done using all the ground truth values of the test set. Although NRMS is a direct metric to assess the performance of the models, it favours predicting the mean of multi-modal distributions rather than one of the possible modes. To assess better the ability of the models to capture the structure of the data, we also use a complementary downstream metric. This metric is more relevant regarding real-world situations as missing data imputation is often used as a preprocessing step to train a model. However, this metric is indirect and results in a less precise evaluation.

As downstream models, we use random forests trained and tested respectively on the imputed train set and the imputed test set. 
The performance of a random forest is measured by the NRMS in the case of a regression and the accuracy in the case of a classification. Because of the randomness of the random forest algorithm, we define the metric as the mean performance of ten random forests trained with different seeds.
We use random forests with 100 estimators and at most 1000 leaf nodes per estimator for computational reasons.

To obtain more stable results from each run, the metrics are averaged on five of the last training steps (see Section \ref{subsec:sota_algo} for details). To account for the randomness of the dataset preprocessing and the one of the models initialisation and training procedure, each experiment is run ten times, using ten different seeds. All compared algorithms are run on the same ten preprocessed datasets to obtain more significant results. The mean of the ten metric measurements obtained is reported in order to compare the algorithms with each other. The sample variance of the ten values obtained is reported for better reproducibility. 

\subsection{Compared Algorithms} \label{subsec:sota_algo}

To validate our approach, we have selected and implemented three state-of-the-art algorithms for comparison: AimNet \cite{Aimnet}, MIDA \cite{MIDA} and MissForest \cite{MissForest}. We have also implemented a custom denoising autoencoder, which we will refer to as DAE.
For DAEMA, we evaluate the model after 39200, 39400, 39600, 39800 and 40000 batch-steps.
For AimNet, we use the hyperparameters described in the paper \cite{Aimnet}. We evaluate the model after 18, 19, 20, 21 and 22 epochs.
For MIDA, we have not found any indication on the batch size. Therefore, we use the whole train set at each training step. Moreover, we use the Adam optimiser with a learning rate of 0.0001 instead of the recommended one as we found it achieves better results. We evaluate the model after 492, 494, 496, 498 and 500 epochs.

We also compare DAEMA with MissForest. 
However, MissForest targets a fix dataset, as it reconstructs the whole dataset at once. Thus it can be seen as a one shot process only, with a computational burden that makes it difficult to apply on a data stream.
To have a fair comparison we had to adapt the algorithm. During the training procedure, we save all predictors produced by each iteration. At test time, each predictor is applied one by one to unseen data.
We use random forests of 100 estimators each without any leaf-node limit. The maximum number of iterations is set to 10. For this approach, as the number of steps is dynamically chosen, we evaluate the model only once, i.e. after convergence or after the maximum number of iterations has been reached.

Finally, we also compare DAEMA against a classical denoising autoencoder (DAE). The training procedure of the DAE is similar to the one of DAEMA. The autoencoder is composed of three fully-connected layers, respectively of size $2d$, $2d$ and $d$, followed by an hyperbolic tangent activation function for the first two layers and no activation function for the last one. We evaluate the model after 39200, 39400, 39600, 39800 and 40000 batch-steps.

\subsection{Comparison with State-of-the-Art Algorithms}
We compare DAEMA with the other algorithms on seven publicly available datasets both in MCAR and MNAR settings (see Section \ref{sec:preprocesing_steps} for dataset preprocessing). The performance of a simple mean imputation \textit{Mean} and a perfect reconstruction \textit{Real} are also reported as indicative lower and upper bounds.
\begin{table*}[t]
\caption{State-of-the-art comparison under the MCAR setting. NRMS is reported in the top table and the performance of a downstream random forest using the imputed datasets is reported in bottom one (accuracy for classification and NRMS for regression).\newline{}}

\begin{subtable}{1.0\textwidth}
\centering
\begin{tabular}{ c c c c c c c c}
\hline
 & \multicolumn{1}{c}{EEG}&\multicolumn{1}{c}{Glass}&\multicolumn{1}{c}{Breast}&\multicolumn{1}{c}{Ionosphere}&\multicolumn{1}{c}{Shuttle}&\multicolumn{1}{c}{Boston}&\multicolumn{1}{c}{CASP}\\
\hline
DAEMA & \textbf{\small 0.392}\tiny~$\pm$.005 & \textbf{\small 0.714}\tiny~$\pm$.093 & \small 0.678\tiny~$\pm$.041 & \textbf{\small 0.745}\tiny~$\pm$.035 & \textbf{\small 0.546}\tiny~$\pm$.123 & \textbf{\small 0.635}\tiny~$\pm$.047 & \textbf{\small 0.422}\tiny~$\pm$.069\\
DAE & \small 0.467\tiny~$\pm$.005 & \small 0.754\tiny~$\pm$.092 & \textbf{\small 0.645}\tiny~$\pm$.037 & \small 0.799\tiny~$\pm$.034 & \small 0.597\tiny~$\pm$.111 & \small 0.645\tiny~$\pm$.057 & \small 0.460\tiny~$\pm$.066\\
AimNet & \small 0.440\tiny~$\pm$.005 & \small 0.926\tiny~$\pm$.108 & \small 0.681\tiny~$\pm$.029 & \small 0.846\tiny~$\pm$.036 & \small 0.568\tiny~$\pm$.123 & \small 0.704\tiny~$\pm$.063 & \small 0.431\tiny~$\pm$.070\\
MIDA & \small 0.925\tiny~$\pm$.015 & \small 0.954\tiny~$\pm$.108 & \small 0.797\tiny~$\pm$.033 & \small 0.901\tiny~$\pm$.020 & \small 1.470\tiny~$\pm$.401 & \small 0.845\tiny~$\pm$.064 & \small 0.861\tiny~$\pm$.051\\
\hline
MissForest & \textbf{\small 0.364}\tiny~$\pm$.006 & \small 0.741\tiny~$\pm$.076 & \small 0.677\tiny~$\pm$.050 & \small 0.756\tiny~$\pm$.040 & \small 0.673\tiny~$\pm$.182 & \textbf{\small 0.601}\tiny~$\pm$.052 & \textbf{\small 0.379}\tiny~$\pm$.077\\
Mean & \small 0.998\tiny~$\pm$.013 & \small 1.003\tiny~$\pm$.112 & \small 0.993\tiny~$\pm$.035 & \small 1.018\tiny~$\pm$.023 & \small 0.969\tiny~$\pm$.077 & \small 0.998\tiny~$\pm$.048 & \small 1.004\tiny~$\pm$.030\\
\end{tabular}
\end{subtable}
\vskip 0.15in

\begin{subtable}{1.0\textwidth}
\centering
\begin{tabular}{c c c c c c|c c}
\multicolumn{1}{c}{ }&\multicolumn{5}{c|}{Accuracy {\small(higher is better)}}&\multicolumn{2}{c}{NRMS {\small(lower is better)}}\\
\hline
 & \multicolumn{1}{c}{EEG}&\multicolumn{1}{c}{Glass}&\multicolumn{1}{c}{Breast}&\multicolumn{1}{c}{Ionosphere}&\multicolumn{1}{c|}{Shuttle}&\multicolumn{1}{c}{Boston}&\multicolumn{1}{c}{CASP}\\
\hline
DAEMA & \textbf{\small 0.861}\tiny~$\pm$.004 & \textbf{\small 0.684}\tiny~$\pm$.043 & \textbf{\small 0.973}\tiny~$\pm$.011 & \small 0.932\tiny~$\pm$.022 & \textbf{\small 0.996}\tiny~$\pm$.001 & \textbf{\small 0.481}\tiny~$\pm$.047 & \textbf{\small 0.708}\tiny~$\pm$.004\\
DAE & \small 0.849\tiny~$\pm$.003 & \small 0.678\tiny~$\pm$.034 & \small 0.971\tiny~$\pm$.011 & \textbf{\small 0.933}\tiny~$\pm$.019 & \small 0.995\tiny~$\pm$.001 & \small 0.492\tiny~$\pm$.049 & \small 0.720\tiny~$\pm$.005\\
AimNet & \small 0.851\tiny~$\pm$.004 & \small 0.669\tiny~$\pm$.043 & \small 0.970\tiny~$\pm$.011 & \small 0.930\tiny~$\pm$.022 & \small 0.996\tiny~$\pm$.001 & \small 0.506\tiny~$\pm$.053 & \small 0.712\tiny~$\pm$.006\\
MIDA & \small 0.823\tiny~$\pm$.005 & \small 0.665\tiny~$\pm$.026 & \small 0.973\tiny~$\pm$.011 & \small 0.930\tiny~$\pm$.024 & \small 0.994\tiny~$\pm$.000 & \small 0.505\tiny~$\pm$.038 & \small 0.761\tiny~$\pm$.004\\
\hline
MissForest & \textbf{\small 0.869}\tiny~$\pm$.006 & \small 0.675\tiny~$\pm$.041 & \small 0.971\tiny~$\pm$.013 & \small 0.931\tiny~$\pm$.016 & \textbf{\small 0.997}\tiny~$\pm$.000 & \small 0.492\tiny~$\pm$.045 & \textbf{\small 0.683}\tiny~$\pm$.005\\
Mean & \small 0.824\tiny~$\pm$.005 & \small 0.669\tiny~$\pm$.044 & \small 0.968\tiny~$\pm$.013 & \small 0.927\tiny~$\pm$.024 & \small 0.996\tiny~$\pm$.000 & \small 0.495\tiny~$\pm$.033 & \small 0.752\tiny~$\pm$.004\\
Real & \small 0.925\tiny~$\pm$.005 & \small 0.750\tiny~$\pm$.040 & \small 0.973\tiny~$\pm$.009 & \small 0.940\tiny~$\pm$.020 & \small 1.000\tiny~$\pm$.000 & \small 0.372\tiny~$\pm$.051 & \small 0.615\tiny~$\pm$.006\\
\end{tabular}
\end{subtable}
\label{tab:sota_mcar}
\end{table*}

\begin{table*}[t]
\caption{State-of-the-art comparison under the MNAR setting. NRMS is reported in the top table and the performance of a downstream random forest using the imputed datasets is reported in bottom one (accuracy for classification and NRMS for regression).\newline{}}
\begin{subtable}{1.0\textwidth}
\centering
\begin{tabular}{ c c c c c c c c}
\hline
 & \multicolumn{1}{c}{EEG}&\multicolumn{1}{c}{Glass}&\multicolumn{1}{c}{Breast}&\multicolumn{1}{c}{Ionosphere}&\multicolumn{1}{c}{Shuttle}&\multicolumn{1}{c}{Boston}&\multicolumn{1}{c}{CASP}\\
\hline
DAEMA & \textbf{\small 0.390}\tiny~$\pm$.006 & \small 0.731\tiny~$\pm$.123 & \small 0.703\tiny~$\pm$.046 & \textbf{\small 0.765}\tiny~$\pm$.080 & \textbf{\small 0.611}\tiny~$\pm$.253 & \textbf{\small 0.634}\tiny~$\pm$.051 & \textbf{\small 0.429}\tiny~$\pm$.052\\
DAE & \small 0.467\tiny~$\pm$.006 & \textbf{\small 0.725}\tiny~$\pm$.129 & \textbf{\small 0.656}\tiny~$\pm$.047 & \small 0.807\tiny~$\pm$.078 & \small 0.683\tiny~$\pm$.217 & \small 0.644\tiny~$\pm$.045 & \small 0.465\tiny~$\pm$.044\\
AimNet & \small 0.440\tiny~$\pm$.008 & \small 0.822\tiny~$\pm$.128 & \small 0.687\tiny~$\pm$.056 & \small 0.853\tiny~$\pm$.072 & \small 0.642\tiny~$\pm$.247 & \small 0.683\tiny~$\pm$.041 & \small 0.435\tiny~$\pm$.053\\
MIDA & \small 0.962\tiny~$\pm$.023 & \small 0.897\tiny~$\pm$.119 & \small 0.820\tiny~$\pm$.069 & \small 0.884\tiny~$\pm$.060 & \small 1.481\tiny~$\pm$.244 & \small 0.842\tiny~$\pm$.033 & \small 0.862\tiny~$\pm$.032\\
\hline
MissForest & \textbf{\small 0.363}\tiny~$\pm$.006 & \small 0.786\tiny~$\pm$.231 & \small 0.688\tiny~$\pm$.051 & \small 0.768\tiny~$\pm$.085 & \small 0.653\tiny~$\pm$.178 & \textbf{\small 0.591}\tiny~$\pm$.059 & \textbf{\small 0.386}\tiny~$\pm$.058\\
Mean & \small 1.041\tiny~$\pm$.024 & \small 0.957\tiny~$\pm$.119 & \small 1.018\tiny~$\pm$.041 & \small 0.994\tiny~$\pm$.057 & \small 1.027\tiny~$\pm$.148 & \small 1.003\tiny~$\pm$.046 & \small 1.013\tiny~$\pm$.030\\
\end{tabular}
\end{subtable}
\vskip 0.15in
\begin{subtable}{1.0\textwidth}
\centering
\begin{tabular}{c c c c c c|c c}
&  \multicolumn{5}{c|}{Accuracy {\small (higher is better)}} & \multicolumn{2}{c}{NRMS {\small (lower is better)}}\\
\hline
 & \multicolumn{1}{c}{EEG}&\multicolumn{1}{c}{Glass}&\multicolumn{1}{c}{Breast}&\multicolumn{1}{c}{Ionosphere}&\multicolumn{1}{c|}{Shuttle}&\multicolumn{1}{c}{Boston}&\multicolumn{1}{c}{CASP}\\
\hline
DAEMA & \textbf{\small 0.870}\tiny~$\pm$.004 & \textbf{\small 0.710}\tiny~$\pm$.037 & \small 0.968\tiny~$\pm$.011 & \small 0.935\tiny~$\pm$.018 & \textbf{\small 0.997}\tiny~$\pm$.001 & \textbf{\small 0.469}\tiny~$\pm$.045 & \textbf{\small 0.696}\tiny~$\pm$.011\\
DAE & \small 0.859\tiny~$\pm$.006 & \small 0.693\tiny~$\pm$.060 & \textbf{\small 0.969}\tiny~$\pm$.013 & \textbf{\small 0.938}\tiny~$\pm$.016 & \small 0.996\tiny~$\pm$.001 & \small 0.479\tiny~$\pm$.048 & \small 0.706\tiny~$\pm$.013\\
AimNet & \small 0.862\tiny~$\pm$.006 & \small 0.681\tiny~$\pm$.048 & \small 0.966\tiny~$\pm$.012 & \small 0.932\tiny~$\pm$.019 & \small 0.997\tiny~$\pm$.001 & \small 0.482\tiny~$\pm$.053 & \small 0.698\tiny~$\pm$.012\\
MIDA & \small 0.837\tiny~$\pm$.009 & \small 0.668\tiny~$\pm$.056 & \small 0.968\tiny~$\pm$.008 & \small 0.932\tiny~$\pm$.020 & \small 0.995\tiny~$\pm$.001 & \small 0.505\tiny~$\pm$.038 & \small 0.739\tiny~$\pm$.018\\
\hline
MissForest & \textbf{\small 0.876}\tiny~$\pm$.005 & \small 0.695\tiny~$\pm$.057 & \small 0.967\tiny~$\pm$.011 & \small 0.930\tiny~$\pm$.021 & \textbf{\small 0.997}\tiny~$\pm$.000 & \textbf{\small 0.462}\tiny~$\pm$.053 & \textbf{\small 0.674}\tiny~$\pm$.009\\
Mean & \small 0.837\tiny~$\pm$.010 & \small 0.672\tiny~$\pm$.034 & \small 0.963\tiny~$\pm$.012 & \small 0.932\tiny~$\pm$.019 & \small 0.997\tiny~$\pm$.001 & \small 0.509\tiny~$\pm$.053 & \small 0.732\tiny~$\pm$.017\\
Real & \small 0.924\tiny~$\pm$.003 & \small 0.748\tiny~$\pm$.039 & \small 0.969\tiny~$\pm$.013 & \small 0.940\tiny~$\pm$.017 & \small 1.000\tiny~$\pm$.000 & \small 0.392\tiny~$\pm$.058 & \small 0.618\tiny~$\pm$.004\\
\end{tabular}
\end{subtable}
\label{tab:sota_mnar}
\end{table*}

As shown in Table \ref{tab:sota_mcar}, DAEMA achieves good results in the MCAR setting both in terms of data reconstruction (top table) and downstream task (bottom table) compared to DAE, AimNet and MIDA. It obtains the best performance for six of the seven datasets, sometimes by a huge margin. DAEMA also performs well in the more challenging MNAR setting as shown in Table \ref{tab:sota_mnar}, which can be explained by the kind of patterns an attention mechanism can learn.

We also compare DAEMA with MissForest. We can see DAEMA is very competitive for both missingness settings. The MNAR setting gives a slight advantage to MissForest though. We hypothesise it is because the local nature of MissForest makes it less sensitive to the introduced bias than DAEMA, which models the data distribution on a global scale. However, the scope of MissForest is limited, as it is meant to reconstruct a fix dataset, while DAEMA can process new data, making DAEMA more suitable for real-world applications.

\subsection{Missingness Sensitivity}
\begin{figure*}[t!]
    \centering
    \begin{subfigure}{0.33\textwidth}
        \centering
        \includegraphics[width=\textwidth]{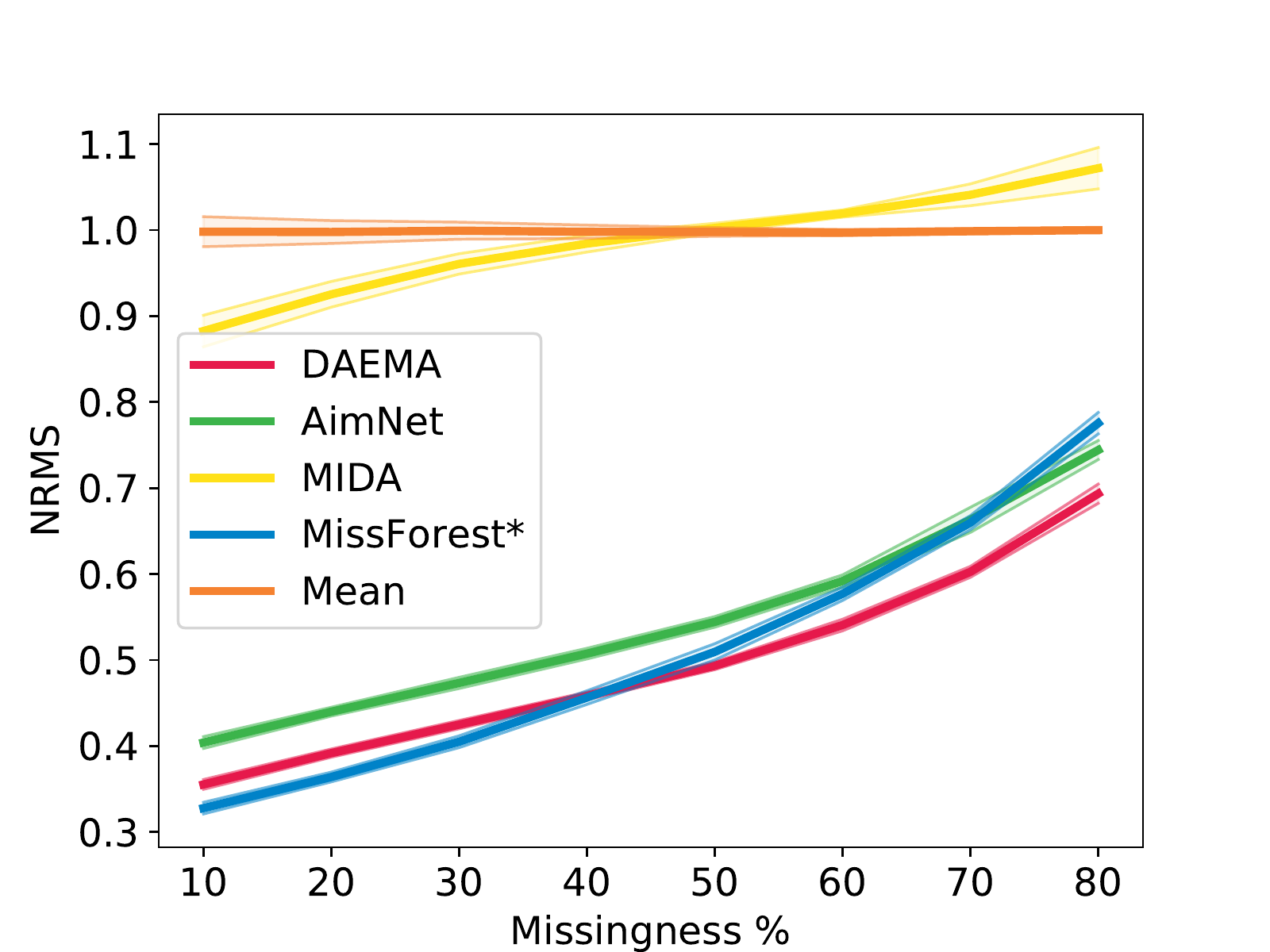}
        \caption{EEG}
        \label{fig:EEG_NRMS}
    \end{subfigure}%
    \begin{subfigure}{0.33\textwidth}
        \centering
        \includegraphics[width=\textwidth]{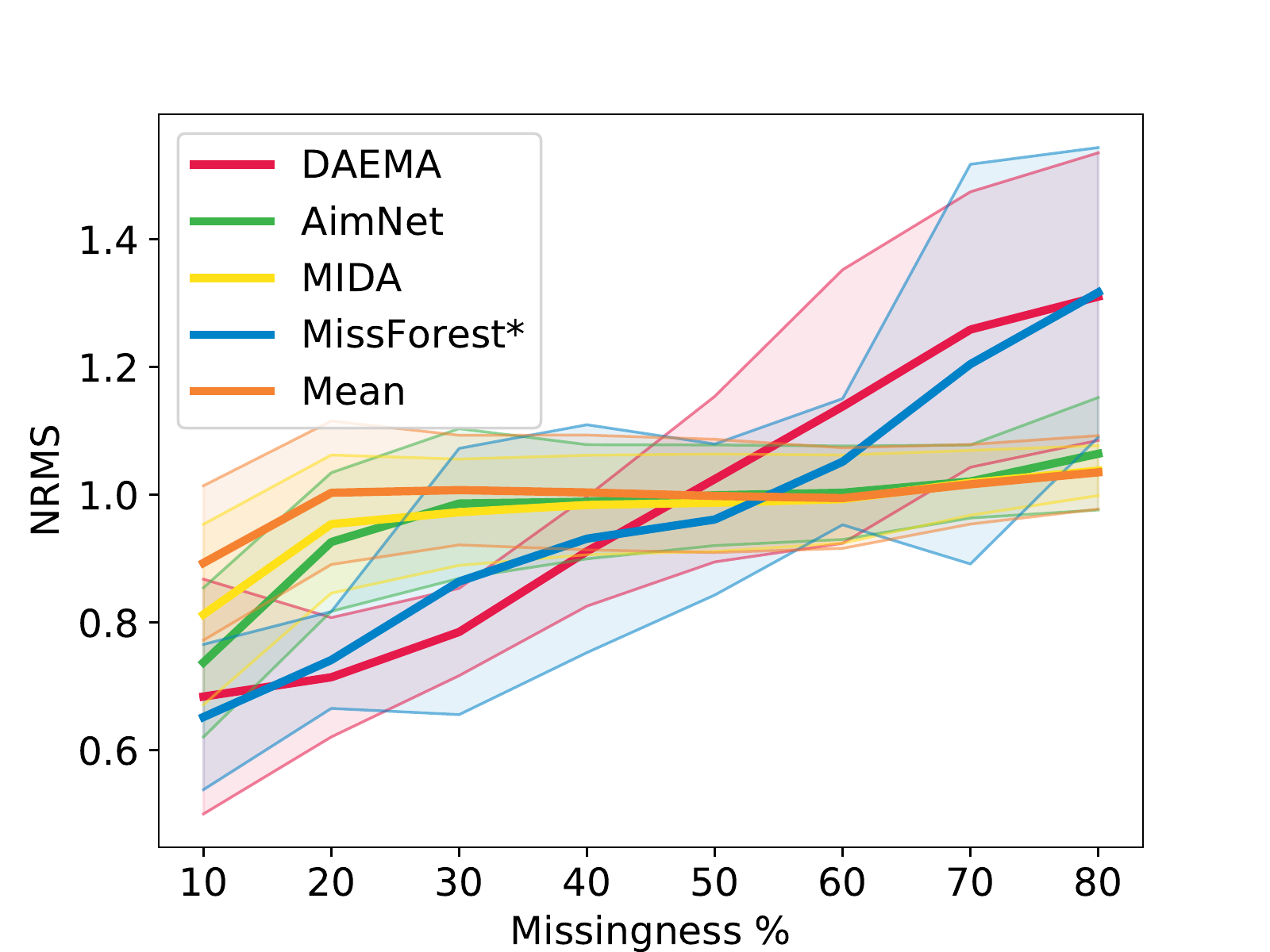}
        \caption{Glass}
        \label{fig:Glass_NRMS}
    \end{subfigure}
    \begin{subfigure}{0.33\textwidth}
        \centering
        \includegraphics[width=\textwidth]{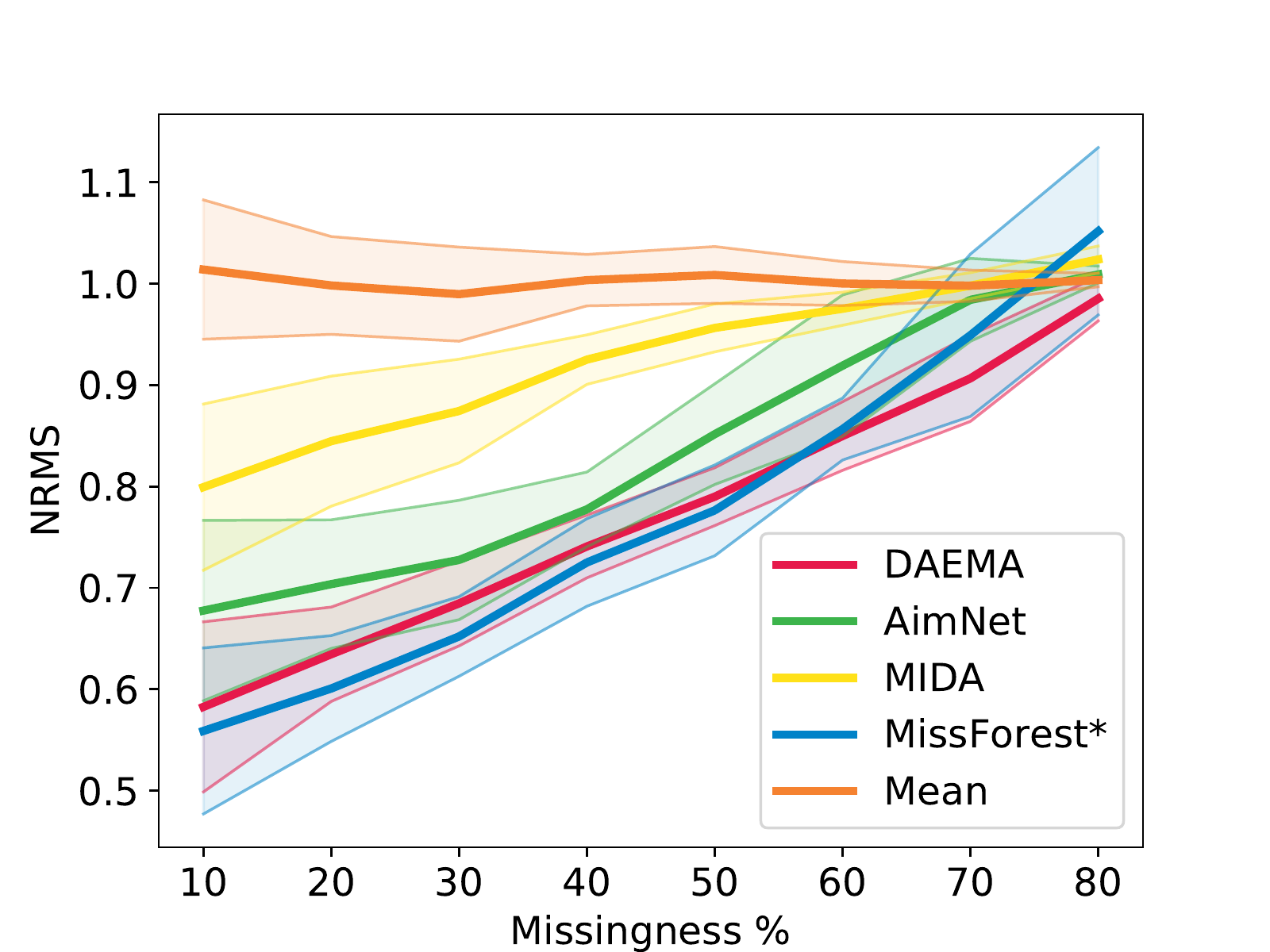}
        \caption{Boston}
        \label{fig:Boston_NRMS}
    \end{subfigure}
    \caption{NRMS reconstruction metric on (\subref{fig:EEG_NRMS}) EEG, (\subref{fig:Glass_NRMS}) Glass and (\subref{fig:Boston_NRMS}) Boston datasets for varying missingness proportions. 
    }
    \label{fig:Missing_NRMS}
\end{figure*}

Figure \ref{fig:Missing_NRMS} shows the sensitivity of the algorithms with respect to the missingness percentage. Mean imputation provides a lower baseline for imputation. We can see that \textit{AimNet}, \textit{MissForest} and \textit{DAEMA} achieve significantly better performance than mean imputation for EEG and Boston datasets. However, their superiority is limited for the Glass dataset. It can be explained by the fact that the Glass dataset has only a few samples (214 samples), showing that \textit{AimNet}, \textit{MissForest} and \textit{DAEMA} need a sufficient amount of data to model the data distribution. This hypothesis is confirmed by the fact that they also seem to be more impacted by the amount of missing data.

Compared to \textit{MissForest}, we can see that \textit{DAEMA} achieves very competitive performance. \textit{DAEMA} seems to be less impacted by the amount of missingness, giving it an advantage for high missingness rates. Furthermore, as explained in Section \ref{subsec:sota_algo}, the scope of \textit{DAEMA} is less limited than the one of \textit{MissForest} making it more adapted to real-world applications.


\section{Conclusion}
In this work, we propose a novel algorithm, DAEMA, a denoising autoencoder with an attention mechanism. Unlike state-of-the-art algorithms that learn from the entire input including placeholder values (mean-imputed or zero-imputed), DAEMA focuses on observed values thanks to its specifically designed attention mechanism. We show that, even when applied to a simple denoising autoencoder without hyperparameter tuning, this new attention mechanism outperforms state-of-the-art approaches on several instances of missing data under \textit{missing completely at random} (MCAR) and \textit{missing not at random} (MNAR) settings. These results also propagate nicely to improve the performance on downstream tasks.



%
%
%
\bibliographystyle{splncs04}
\bibliography{references.bib}

\end{document}